\documentclass[twoside,11pt]{article}

\usepackage{booktabs}
\usepackage{listings}
\usepackage{amsmath}

\usepackage{jmlr2e}

\ShortHeadings{Augmentor}{Bloice, Stocker, and Holzinger}
\firstpageno{1}

\begin{document}

\title{Augmentor: An Image Augmentation Library for Machine Learning}

\author{\name Marcus D. Bloice \email marcus.bloice@medunigraz.at
       \AND
       \name Christof Stocker \email stocker.christof@gmail.com
       \AND
       \name Andreas Holzinger \email andreas.holzinger@medunigraz.at \\
       \addr Institute for Medical Informatics, Statistics, and Documentation\\
       Medical University of Graz\\
       Graz, Austria}

\maketitle

\begin{abstract}
The generation of artificial data based on existing observations,
known as data augmentation, is a technique used in machine learning to improve 
model accuracy, generalisation, and to control overfitting.
Augmentor is a software package, 
available in both Python and Julia versions, that provides a high level
API for the expansion of image data using a stochastic, pipeline-based approach which effectively 
allows for images to be sampled from a distribution of augmented images at runtime. 
Augmentor provides methods for most standard augmentation practices as well as 
several advanced features such as label-preserving, randomised elastic distortions, and provides
many helper functions for typical augmentation tasks used in machine learning.
\end{abstract}


\section{Introduction}

Data augmentation is the artificial generation of
data through the introduction of new samples created
by the perturbation of the original dataset, while preserving the label 
of newly generated samples.
It is a convenient and frequently
employed method for generating more training data at low effort,
or when the accumulation of new samples is no longer feasible,
such as in a discontinued clinical trial.
Data augmentation is most commonly utilised in the branch of
machine learning that concerns image analysis \citep{hauberg2015}.

The Augmentor project uses a stochastic, pipeline-based approach 
to image augmentation. The pipeline approach allows the user to 
chain augmentation operations together, such as shears, rotations, and crops, 
and pass images through this 
pipeline in order to create new data. All operations in the pipeline 
are applied stochastically, both in terms of the probability of the 
operations being applied to each image as the image passes through the 
pipeline, and in terms of each operation's parameters, which are also 
randomised within user specified ranges. This effectively allows you 
to sample from a distribution of possible images, generated by the 
pipeline at runtime.

Therefore, the aim of the package is to provide a comprehensive 
and highly customisable image augmentation library, which is platform 
independent but also independent from any particular machine
learning framework. Crucial to the successful application of augmentation is the generation of 
realistically feasible training data, meaning tight control of the pipeline
is a necessity when creating new data. Augmentor's operations are therefore
highly parametric, allowing fine control over how images are created.

\section{Documentation and Availability}
The Augmentor package is available for Python and Julia. Sources are 
available on GitHub, while comprehensive documentation is hosted on Read The Docs (see Table~\ref{tbl:availability}). Both versions of the Augmentor package are available under the terms of the MIT Licence. 

\begin{table}[h!]
\centering
\caption{Augmentor availability.}
\label{tbl:availability}
\begin{tabular}{@{}lll@{}}
\toprule
				  & \textbf{Source}                         & \textbf{Documentation}             \\ \midrule
\textbf{Python}   & https://github.com/mdbloice/Augmentor   & http://augmentor.readthedocs.io    \\
\textbf{Julia}    & https://github.com/Evizero/Augmentor.jl & http://augmentorjl.readthedocs.io  \\ \bottomrule
\end{tabular}
\end{table}

To install Augmentor:

\begin{itemize}
\setlength\itemsep{0.1em}
\item Python: \texttt{pip install Augmentor}
\item Julia: \texttt{Pkg.clone("https://github.com/Evizero/Augmentor.jl.git")}
\end{itemize}

\section{Design}
\label{sec:design}
We took into account typical augmentation techniques from the literature, and techniques reported on various competition sites such as Kaggle, when developing the API. Standard operations include arbitrary rotations, transformations through the horizontal and vertical axes, cropping, scaling, perspective shifting, shearing, and zooming \citep{dosovitskiy2013,simard2003,krizhevsky2012,howard2013}. Less frequently used operations were also implemented \citep{dosovitskiy2015}, as well a number of pre-processing techniques in common use. Also, a large number of convenience functions have been implemented that take into account typical augmentation techniques.

\begin{figure}[h!]
    \centering
    \includegraphics[scale=0.45]{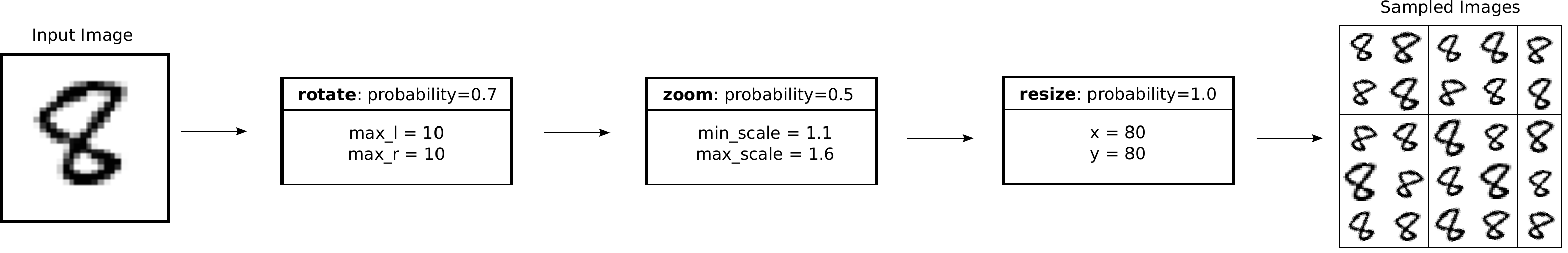}
    \caption{An example pipeline, with three operations. From a single image on the left, variants can be generated by passing the image multiple times through the pipeline. Each time the image is passed through, operations are either applied or skipped based on the user-defined probability parameter. If an operation is applied, the parameters of the operation itself are chosen randomly within a user specified range---for example, the rotate operation's rotation angle is chosen at random from between $-10^{\circ}$ and $+10^{\circ}$.}
    \label{fig:pipeline}
\end{figure}

Because image augmentation is often performed accumulatively, a pipeline-based API was developed (see Figure~\ref{fig:pipeline}). To use Augmentor, you begin with an empty pipeline. The user adds operations to this pipeline in the order they wish the operations to be applied to images that are passed through the pipeline. As well as this, the user can specify the probability that each operation should be applied to images as they pass through.
Also, the range of each operation's freedom of movement is likewise defined by the user, for example by specifying that a rotation operation can operate within the range of $-10^{\circ}$ to $10^{\circ}$. Once a pipeline has been generated, an image or set of images are repeatedly passed though the pipeline until the desired amount of new images have been generated. The stochastic nature of the pipeline approach will produce different image data each time an image passed through the pipeline. This stochastic approach allows for a potentially very large amount of images to be generated from even a small initial dataset.

\subsection{Main Features}
A complete list of features can be found in the project's documentation. 
Some commonly used features are random rotations, transforms through the horizontal 
and vertical axes, cropping (randomly positioned or centred), random zoom levels, 
random scaling and resizing. 
Other transforms, such as shearing by random angles in random directions and through random axes, 
as well as perspective transformations are also implemented. Operations have been implemented with machine learning in mind. For example, arbitrary rotations will not result in images with black or transparent regions around the newly rotated image, as the images are optimally cropped and then resized to their original input size. The same is true of the shear and perspective tilt operations.

\begin{figure}[h!]
    \centering
    \includegraphics[scale=0.3]{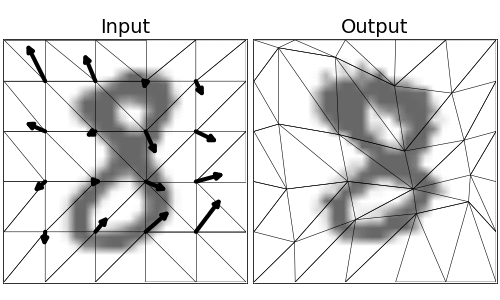}
    \caption{Randomised, label-preserving elastic distortions.}
    \label{fig:elastic}
\end{figure}

Augmentor also has the capability of producing random 
elastic transforms \citep{simard2003}, and to perform these 
in a highly configurable way. The user may specify a grid size which 
controls the granularity of the distortions and the strength of the 
displacement within the grid (the magnitude of the arrows shown in Figure~\ref{fig:elastic}).

\subsection{Practical Example}
To demonstrate the API and to highlight the effectiveness of augmentation on a well-known dataset, a short experiment was performed. Using the MNIST dataset, a CNN was trained on 1000 random images (100 samples per class) extracted from the 60,000 image training set and tested on the standard 10,000 image test set. Then this set of 1000 images was augmented to produce 10,000 new images, a separate CNN was trained using the augmented dataset, and the results of the models were compared. As shown in Table~\ref{tbl:results} this resulted in an almost 4\% improvement in performance on the same test set.

This augmentation experiment was perform using randomised elastic transforms and random rotations. To show how the Augmentor API works in practice, we will demonstrate how this augmented dataset was generated. To begin, a pipeline object is created, pointing to a folder containing the images:

\begin{lstlisting}
import Augmentor
p = Augmentor.Pipeline("/path/to/mnist/1")
\end{lstlisting}

Now that a pipeline object, \texttt{p}, has been created, operations are added to the pipeline as follows: 

\begin{lstlisting}
p.random_distortion(probability=1, grid_width=4, grid_height=4, magnitude=5)
p.rotate(probability=0.5, max_left_rotation=10, max_right_rotation=10)
\end{lstlisting}

Every operation has at least a probability parameter. This was set to 1.0 for the randomised distortions and 0.5 for the rotation operation. Finally, the sample function is called to generate the data:

\begin{lstlisting}
p.sample(1000)
\end{lstlisting}

This generates 1000 new augmented images. The procedure was repeated 10 times, once per digit, for 10,000 augmented images in total.

\begin{table}[]
\centering
\caption{Augmentation Example}
\label{tbl:results}
\begin{tabular}{@{}lll@{}}
\toprule
Experiment & Dataset                             & Test Set Accuracy    \\ \midrule
Baseline   & 1,000 image training set            & 93.94\%              \\
Augmented  & 11,000 image augmented training set & 97.28\%              \\ \bottomrule
\end{tabular}
\end{table}

\section{Conclusions}
Image augmentation is an important constituent of many machine learning tasks, 
particularly deep learning. Augmentor makes it easier to perform artificial data generation, by providing a stochastic, pipeline-based API that allows for fine-grained control over the creation of augmented data and provides many functions for augmentation techniques found in the literature. Future work will entail expanding functionality, such as the ability to mirror augmentation on a reference data set, or mimicking more advanced preprocessing and augmentation methods such as the specialised contrast manipulation techniques or vignetting shown in \cite{wu2015}.

\bibliography{augmentor}

\end{document}